\documentclass[letterpaper]{article} 
\usepackage[]{aaai25}  
\usepackage{times}  
\usepackage{helvet}  
\usepackage{courier}  
\usepackage[hyphens]{url}  
\usepackage{graphicx} 
\usepackage{natbib}  
\usepackage{caption} 
\usepackage{algorithm}
\usepackage{algorithmic}
\usepackage{amsfonts}
\usepackage{amsmath}

\usepackage{newfloat}
\usepackage{listings}
\DeclareCaptionStyle{ruled}{labelfont=normalfont,labelsep=colon,strut=off} 
\floatstyle{ruled}
\newfloat{listing}{tb}{lst}{}
\floatname{listing}{Listing}
\title{Learning Multi-Agent Multi-Machine Tending by Mobile Robots}
\author{
    Abdalwhab Abdalwhab\textsuperscript{\rm 1,2},
    Giovanni Beltrame\textsuperscript{\rm 3},
    Samira Ebrahimi Kahou\textsuperscript{\rm 2,4},
    David St-Onge\textsuperscript{\rm 1}
    \thanks{We thank NSERC USRA and Discovery programs for their financial support. We also acknowledge the support provided by Calcul Québec, Compute Canada,  CIFAR, and Google.}
}
\affiliations{
    \textsuperscript{\rm 1}The Department of Mechanical Engineering, ETS Montreal\\
    \textsuperscript{\rm 2}Mila - Quebec Artificial Intelligence Institute\\
    \textsuperscript{\rm 3}Department of Computer and Software Engineering, Polytechnique Montréal\\
    \textsuperscript{\rm 4}University of Calgary and Canada CIFAR AI Chair\\
    
    Abdalwhab.Bakheet@gmail.com
}

\usepackage{bibentry}

\begin{document}

\maketitle

\begin{abstract}

Robotics can help address the growing worker shortage challenge of the manufacturing industry. As such, machine tending is a task collaborative robots can tackle that can also highly boost productivity. Nevertheless, existing robotics systems deployed in that sector rely on a fixed single-arm setup, whereas mobile robots can provide more flexibility and scalability. We introduce a multi-agent multi-machine tending learning framework by mobile robots based on Multi-agent Reinforcement Learning (MARL) techniques with the design of a suitable observation and reward. Moreover, we integrate an attention-based encoding mechanism into the Multi-agent Proximal Policy Optimization (MAPPO) algorithm to boost its performance for machine tending scenarios. Our model (AB-MAPPO) outperforms MAPPO in this new challenging scenario in terms of task success, safety, and resource utilization. Furthermore, we provided an extensive ablation study to support our design decisions. 

\end{abstract}

%

\section{Introduction}
The manufacturing sector increasingly relies on robotics to address workforce shortages and enhance efficiency \cite{Logan2022laborshortage}, particularly in repetitive and labor-intensive tasks. Traditionally, industrial robotics involves fixed robotic systems. A typical task is machine tending that automates the loading and unloading of materials and parts from production machines \cite{heimann2023mobile}. While effective, achieving this task with fixed setups is limited in flexibility and demands considerable human oversight.


Instead of a fixed robotic arm dedicated to a single machine, a fleet of mobile manipulators could autonomously navigate between multiple machines and storage areas, significantly boosting productivity and machine utilization.

However, the deployment of multi-robot systems introduces complex challenges in coordination and control. Predominantly, solutions in the industry, such as those implemented in Amazon's warehouses, have used a centralized approach where a single server orchestrates the tasks, trajectory planning, and collision avoidance of all robots, ensuring optimized fleet operations. This centralized model, while efficient, suffers from several disadvantages including single points of failure, high reliance on continuous communication, and limited adaptability in dynamic or unstructured environments \cite{fan2018fully}.

In contrast, decentralized robotic systems, where each unit operates independently and makes decisions in real-time, offer robust alternatives free from the constraints of centralized control. Reinforcement learning (RL), particularly with its recent advances in deep learning and computational capabilities supported by GPUs, presents a promising framework for such autonomous decision-making. RL enables robots to learn optimal behaviors through trial and error, using feedback from their environment to refine their actions \cite{singh2022reinforcement}. 

Despite being a very promising technique, most of the works that investigate RL for multi-agent tasks, do it for games like Starcraft Multi-Agent Challenge (SMAC) \cite{guo2024heterogeneous,zhou2021cooperative} and Soccer \cite{lobos2022ma} or simplified scenarios such as goal coverage and formations \cite{agarwal2019learning, long2020evolutionary,chen2020distributed} distancing MARL from deployable real-world scenarios. 


In contrast, this paper introduces a novel reinforcement learning model tailored to a realistic scenario of decentralized management of mobile robots for machine tending in manufacturing settings autonomously handling both task assignment and navigation. Aiming to leverage the full potential of autonomous systems in enhancing operational efficiency.  In short, our main contributions are: 1)Propose and formulate a new more realistic, and challenging scenario for multi-agent navigation for machine tending; 2) Design a novel dense reward function specifically for this scenario; 3) Design an attention-based encoding technique and incorporate it into a working baseline (AB-MAPPO). We also include an ablation study to verify the effect of various design decisions.

\section{Background and Related Work}\label{sec:RelatedWorks}

\subsection{Machine Tending}

Despite the substantial research and development efforts in both academia and industry \cite{li2023assistive}, the gap between automation demands and the available commercial products is still wide and needs further work to be reduced.

Due to the complexity of the machine tending problem, different researchers focused on different specific areas. For instance, Jia et al.~\cite{jia2022intelligent} focused on recognizing the state of the machine by first detecting a specific computer numerical control (CNC) machine and then using text detection and recognition techniques to get the status from its display.

Al-Hussaini et al.~\cite{al2020human} developed a semi-autonomous mobile manipulator that given a task from the operator generates the execution plan. It also estimates the risk of the plan execution and provides a simulation of the plan for the operator to approve it to be executed autonomously or to teleoperate the robot. In addition, Heimann et al.~\cite{heimann2023mobile} presented a ROS2-based mobile manipulator solution with a focus on evaluating the system precision in reaching the machine location and reaching inside the machine. Moreover, Chen et al.~\cite{chen2021integrated} proposed a centralized method for multi-agent task assignment and path planning, where task assignment is informed by the task completion cost. Behbahani et al.~\cite{behbahani2021episodic} used a learning-from-demonstration approach to teach a physical fixed robot to tend to one machine in a simplified tabletop setup. 

In a different direction, Burgess-Limerick et al.~\cite{burgess2024reactive} developed a controller to allow a mobile manipulator to perform pick and place task on the move in a dynamic environment, reducing the overall task execution time.

\subsection{Reinforcement Learning}




RL has been explored for various robotic tasks such as pick and place, navigation, collision avoidance, locomotion, and quadcopter control. \cite{singh2022reinforcement,zhu_deep_2021,wu2023daydreamer}. Generally, RL methods can be divided into two main categories, on one hand, model-free techniques try to directly learn a policy that maximizes the returned reward. For instance, Proximal Policy Optimization (PPO) \cite{schulman2017proximal} is a very famous and widely used example, along with Deep Q-Networks (DQN) \cite{mnih2013playing} and actor-critic methods \cite{konda1999actor}. On the other hand, model-based techniques try to explicitly learn a model for the environment dynamics and use that model to plan the actions directly \cite{hafner2019learning}, or to learn a policy by interaction with the learned dynamics model to reduce the needed interactions with the environment \cite{sutton1991dyna, wu2023daydreamer}.

However, RL for machine tending is not a well-explored area of research, except for some works such as ~\cite{iriondo2019pick}, where they used PPO and DDPG to teach a single mobile manipulator to navigate to the location of a table that has the object to pick. Despite targeting one agent and one machine setup, their work is still far from being deployable in reality because it does not address part dropping, learning to navigate when the part is ready, or navigating multiple times to pick and place multiple parts.   

On the other hand, multi-agent reinforcement learning (MARL) is a growing research area that focuses on extending RL to multi-agent setups to expand its potential use. Two famous works in this area are MAPPO \cite{yu2021surprising} and MADDPG \cite{lowe2017multi} that extended PPO and Deep Deterministic Policy Gradient (DDPG) respectively.


To the best of our knowledge, one of the only works presenting a MARL solution for machine tending is \cite{agrawal2021multi}. This pioneering work introduces a multi-agent framework for integrated job scheduling and navigation for shop floor management exhibiting robustness under processing delays and failures. However, their implementation assumes the agents can accumulate parts from various machines without unloading them,  as if they had infinite onboard storage, making the solution non-deployable in the real world. 

Distinct from prior models, our work addresses multi-agent, multi-machine tending by encompassing task assignment, the navigation to machines with ready parts, and the transport of these parts to designated storage areas. This approach more realistically reflects real-world manufacturing scenarios, pushing MARL closer to full deployment on machine tending tasks. Moreover, our work can be integrated with low-level controllers like ~\cite{burgess2024reactive} to facilitate actual deployment in real robots.

\section{Problem Definition}\label{sec:ProblemDefinition}
The multi-agent, multi-machine tending problem involves coordination, navigation, and object manipulation, where N agents service M machines, managing tasks from feeding raw materials to delivering finished parts to storage. The problem's constraints include:

\begin{itemize}
\item Coordination: Agents autonomously determine their targets to optimize machine usage and minimize collision risks.

\item Navigation: Agents must navigate efficiently to and from machines, avoiding collisions with other agents and static obstacles while transporting parts to the storage area.

\item Cooperative-Competitive Dynamics: Agents compete for parts; the first to arrive at a machine can claim the part, requiring others to wait for the next availability.

\item Temporal Reasoning: Agents must sequence their actions correctly, ensuring machines are fed or parts are collected and delivered at appropriate times.
\end{itemize}

Object manipulation is acknowledged but not addressed in this study, focusing on the four constraints above. Key assumptions include instantaneous pick and place actions and external management of raw material supply, as supported by prior studies \cite{burgess2024reactive}.

\section{Methodology}\label{sec:Methodology}
\subsection{MAPPO backbone}
MAPPO, a prominent multi-agent reinforcement learning algorithm introduced by Yu et al.~\cite{yu2021surprising}, operates on a dual-network architecture comprising an actor and a critic. The actor network, or policy, executes actions aimed at maximizing accumulated discounted returns, which are estimated by the critic based on the current state values.

This algorithm is designed for a decentralized partially observable Markov decision process (DEC-POMDP), characterized by the number of agents ($n$), the global state space ($\mathcal{S}$), the local observation space for agents ($\mathcal{O}$), with each agent $i$ having a local observation $o_i$ derived from the global state $s$,
the action space ($\mathcal{A}$), where each agent acts according to a policy $\pi_\theta(o_i)$, with $\theta$ representing the learnable parameters of the policy, the probability of transitioning from state $s$ to state $\acute{s}$ given the joint actions $A = (a_1, a_2, ..., a_n)$ of the agents ($\mathcal{P}(\acute{s}|s,A)$), and the shared reward function for the joint actions of the agents ($R(s,A)$). MAPPO's objective is to maximize the expected sum of discounted rewards, formulated as:
\begin{equation}
J(\theta) = \mathbb{E}_{A^t,s^t}[\sum_t \gamma^t R(s^t,A^t)]
\end{equation}
where $\gamma$ is the discount factor that prioritizes immediate rewards over future ones.

The algorithm employs a centralized training but decentralized execution (CTDE) approach, where the critic, accessible only during training, utilizes global information across all agents. In contrast, the actor operates solely on local observations during both training and execution phases.

In our implementation, given that our agents are homogeneous, we adopt parameter-sharing to streamline the learning process. However, our scenario deviates from standard MAPPO applications as it encompasses both cooperative and competitive elements. Thus, we have explored strategies involving both individual rewards and shared rewards to accommodate the competitive dynamics among agents.

\subsection{Novel Attention-based Encoding for MAPPO }\label{subsec:Attention}
\begin{figure}[t]
    \centering
    \includegraphics[width=0.85\linewidth]{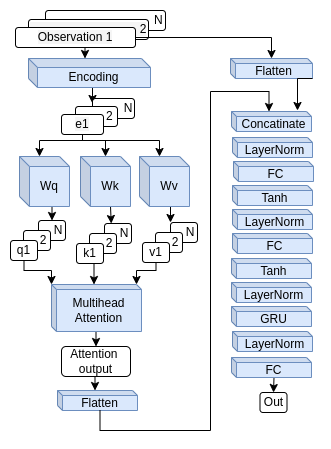}
    \caption{Our Attention-based encoding for the critic}
    \label{fig:MachineTendingArc}
\end{figure}

We introduce AB-MAPPO: an upgraded MAPPO's architecture with the integration of a novel attention-based encoding mechanism into the critic network. This enhancement allows the actor to make decisions based on more accurately estimated values from the critic, thereby improving its performance indirectly as well. Figure \ref{fig:MachineTendingArc} illustrates the design of our attention-based encoding mechanism for the critic. This design enables the critic to leverage all agents' observations to develop a representation that encapsulates both spatial and temporal dynamics effectively.  

Spatial information is captured through encoding and multi-head attention mechanisms. Temporal information is processed using a Gated Recurrent Unit (GRU). Initially, each agent's observation $o_i$ is transformed into an encoded vector $e_i$ using shared learnable parameters $W_e$ and $b_e$, $e_i = W_e * o_i + b_e$.

Each encoded vector $e_i$ is then mapped to three vectors: query $q_i$, key $k_i$, and value $v_i$, using parameters $W_q$, $W_k$, and $W_v$, respectively:
\begin{align}
    q_i & = W_q * e_i \\
    k_i & = W_k * e_i \\
    v_i & = W_v * e_i
\end{align}

These vectors are subsequently processed by a multi-head attention module, which learns the inter-agent attention dynamics:
\begin{align}
    Q & = stack(q_1,q_2,...,q_N) \\
    K & = stack(k_1,k_2,...,k_N) \\
    V & = stack(v_1,v_2,...,v_N) 
\end{align}

The output from the multi-head attention module, based on the seminal paper of Vaswani et al. \cite{vaswani2017attention}, is computed as follows:
\begin{gather}
    M_{out} = Concat(head_1, ..., head_H)W^o \\
    head_j = Attention(QW^Q_j, KW^K_j, VW^V_j) \\
    Attention(Q,K,V) = softmax(QK^T/ \sqrt{d_k}) V , 
\end{gather}
where $W^o, W^Q, W^K$, and $W^V$ are learnable parameters, $d_k$ denotes the key vector dimension, and $H$ represents the number of heads used (in this case, three).

The attention module's output is then flattened and concatenated with the original observations and passed through a sequence of fully connected layers (FCs), normalization (Norm), a GRU, and a final FC layer to produce the value estimate. This layer configuration follows a standard, open-source implementation of MAPPO. For the actor, we retained the standard configuration, which includes a similar series of FCs, Norms, and a GRU layer, differing only in the final output layer, which matches the number of possible actions.

\subsection{Observation Design}\label{subsec:Observation}
In addressing the multi-agent, multi-machine tending problem, designing an efficient observation scheme that captures essential information while excluding superfluous details was crucial. After extensive empirical testing to determine the optimal data inclusion by trying different observation options (more about it in section \ref{sec:AblationStudy}), each agent's observation was crafted to include:

\begin{itemize}
    \item The agent's absolute position and a "has part" flag indicating whether it has picked up a part but not yet placed it.
    \item The relative positions of each machine, accompanied by a flag indicating the readiness of a part at each location.
    \item The relative position of the storage area.
    \item The relative positions of other agents, along with their respective "has part" flags.
\end{itemize}

All positional data represent the central point of entities, normalized to scale values between 0 and 1 to facilitate processing.

\subsection{Reward Design}
The reward structure was tailored to enhance navigation and efficiency in machine-tending tasks:

\noindent\textbf{1. Base Rewards:}

\begin{itemize}
\item Pick Reward ($R_{pi}$): Granted when an agent without a part reaches a machine with a ready part.
\item Place Reward ($R_{pl}$): Awarded when an agent with a part reaches the storage area to place the part.
\item Collision Penalty ($R_c$): Incurred upon collision with agents, walls, machines, or the storage area.
\end{itemize}

\noindent\textbf{2. Distance-based Rewards:}

\begin{itemize}
\item Progress towards the closest machine with a ready part ($R_{pm}$) and the storage area ($R_{ps}$) is rewarded to provide continuous feedback. The rewards are calculated based on the reduction in distance to the target ($d_m^t$) between consecutive steps, scaled by a factor ($pr$). 

\begin{equation}\label{equ:Rpm}
    R_{pm}^t = \begin{cases}
      pr * (d_m^{t-1} - d_m^t) & \text{if the agent has no part}\\
      0 & \text{otherwise}
    \end{cases}   
\end{equation}

\begin{equation}\label{equ:Rps}
    R_{ps}^t = \begin{cases}
      pr * (d_s^{t-1} - d_s^t) & \text{if the agent has part}\\
      0 & \text{otherwise}
    \end{cases}   
\end{equation}
\end{itemize}

\noindent\textbf{3. Utilization Penalty ($R_u$):}

\begin{itemize}
\item To discourage idle machines, a penalty is applied for uncollected parts ($p_{un}$) at each step, calculated based on the number of steps a part (i) remains uncollected ($ns_{i}$) multiplied by a scaling factor ($u$).
\begin{equation}\label{equ:RU}
    R_u = \begin{cases}
       p_{un} * u & \text{if fixed uncollected penalty }\\
      \sum{ns_{i}} * u & \text{otherwise}
    \end{cases}   
\end{equation}
\end{itemize}

\noindent\textbf{4. Time Penalty ($R_t$):}

\begin{itemize}
\item Imposed to encourage movement, particularly when no other rewards or penalties are being applied.

\begin{equation}\label{equ:Rtime}
    R_t = \begin{cases}
      tp & \text{if $R_{pi}$, $R_{pl}$, and $R_u$ are 0 }\\
      0 & \text{otherwise}
    \end{cases}   
\end{equation}
\end{itemize}

The cumulative reward for an agent is the sum of all the above components:
\begin{equation}\label{equ:RT}
    R_T = R_{pi} + R_{pl} + R_c + R_{pm} + R_{ps} + R_u + R_t
\end{equation}

\section{Experiments}\label{sec:Experiments}

\subsection{Simulation Setup}
For our experiments, we leveraged the Vectorized Multi-Agent Simulator (VMAS) \cite{bettini2022vmas}, specifically designed for efficient multi-agent reinforcement learning (MARL) benchmarking. VMAS integrates a 2D physics engine and supports the well-known Gym simulation environments interface. Its architecture allows for straightforward customization and expansion, facilitating the development of new and more complex scenarios.

\begin{figure}[t]
  \centering
  \includegraphics[width=.99\columnwidth]{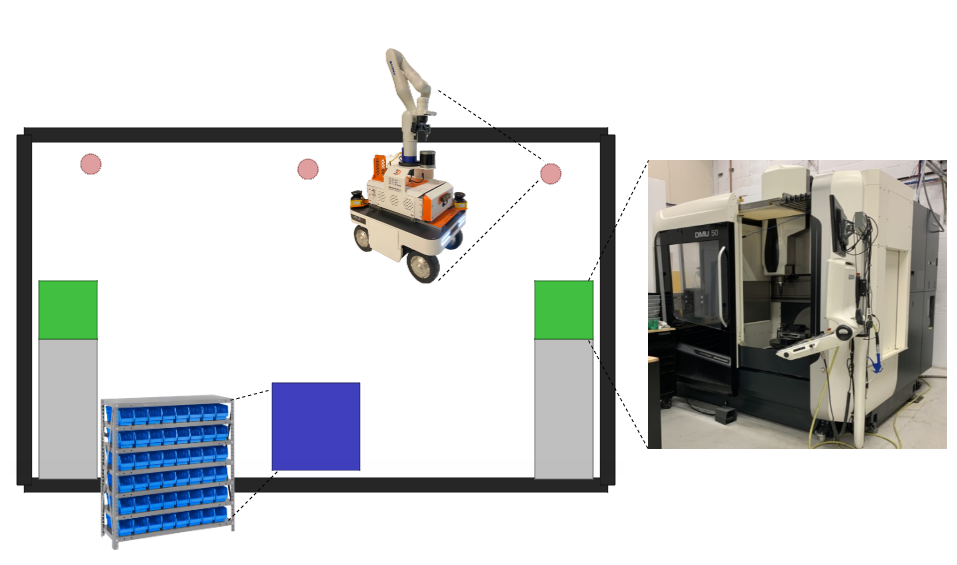}
  \caption{Multi-agent (red) multi-machine (green) tending scenario designed in VMAS, including obstacles (gray) and storage area (blue), with dotted lines pointing to the actual robot (our RanGen robot composed of a Kinova Gen3 arm on top of an AgileX Ranger Mini mobile base), machines (CNC Universal Milling Machine DMU 50) and storage shelves that we are planning to use for real deployment.}
  \label{fig:our_machine_tending}
\end{figure}

Figure \ref{fig:our_machine_tending} illustrates the multi-agent multi-machine tending scenario we developed. In this setup, three agents, depicted in red, initiate from their designated idle positions at the top of the environment. The production machines are highlighted in green, with machine blockers in gray obstructing direct access to the machines from one side. The storage area is marked in blue, and the walls outlining the environment are shown in black. While dotted lines in the figure point to the actual robot(our RanGen robot composed of a Kinova Gen3 arm on top of an AgileX Ranger Mini mobile base), machines (CNC Universal Milling Machine DMU 50) and storage shelves that we are planning to use for real deployment.

In VMAS agents' actions are represented as physical forces $f_x$ and $f_y$. The physics engine takes into account this input force and other factors like the velocity damping, gravity, and forces due to the collisions with other objects \cite{bettini2022vmas}. We used 5 discrete actions 0: no force (translated internally to $f_x=0, f_y=0$), 1: Leftward acceleration ($f_x=-1,f_y=0$), 2: rightward acceleration ($f_x=1,f_y=0$), 3: downward acceleration ($f_x=0,f_y=-1$) and 4: upward acceleration ($f_x=0,f_y=1$). This holonomic motion is also supported by our intended mobile base (the AgileX Ranger Mini).

Each episode in our simulation environment spans 200 time steps, during which agents aim to collect and deliver as many parts as possible. Agents are restricted to carrying only one part at a time; an agent must deliver its current part before picking up another. Additionally, once a part is collected from a machine, there is a production delay of 20 steps before the next part is available from the same machine.

\subsection{Evaluation Procedure}
Our evaluation framework is designed to assess both task effectiveness and resource efficiency through various criteria:

\noindent\textbf{1. Task Success Factors:}

\begin{itemize}
\item Total Number of Collected Parts: Measures the efficiency of part collection by agents.
\item Total Number of Delivered Parts: Assesses the effectiveness of the delivery process.
\end{itemize}

\noindent\textbf{2. Safety Factor:}

\begin{itemize}
\item Total Number of Collisions: Calculated as the cumulative collisions across all agents, indicating the safety of navigation and interaction.
\end{itemize}

\noindent\textbf{3. Resource Utilization Factors:}

\begin{itemize}
\item Machine Utilization (MU): 
\begin{equation}\label{equ:MU}
    MU_i = \frac {P_i} {P_{imax}}, Avr(MU) = \frac {\sum{MU_i}} {M} 
\end{equation}

Where $MU_i$ represents the utilization rate of machine $i$, calculated by dividing the number of parts collected from machine $i$ ($P_i$) by the maximum parts it could have produced ($P_{imax}$). The average utilization across all machines ($Avr(MU)$) is then determined by averaging the $MU_i$ values for all machines ($M$).

\item Agent Utilization (AU): 

\begin{equation}\label{equ:AU}
    AU_i = \frac {P_i} {(P_{max}/N)}, Avr(AU) = \frac{\sum{AU_i}}{N} 
\end{equation}

Where $AU_i$ is the utilization for agent $i$, based on the number of parts it collected ($P_i$) relative to the average expected parts per agent ($P_{max}/N$). The average utilization for all agents ($Avr(AU)$) is the mean of $AU_i$ values across all agents ($N$).

\end{itemize}

Each model's interaction with the environment spanned 18,200 episodes, with performance metrics averaged over the final 200 episodes to assess stability and effectiveness. The experiments were replicated three times using different random seeds to ensure reliability, with results reported as both average values and standard deviations.

\subsection{Results and Adaptability}

Table \ref{tab:ComparisonResults} presents a comparative analysis of AB-MAPPO against the standard MAPPO. The data clearly demonstrate that AB-MAPPO significantly surpasses MAPPO across all evaluated metrics. Notably, it achieves a 40\% reduction in collisions, enhancing safety. Furthermore, it exhibits a 16\% increase in parts collection and a 20\% improvement in parts delivery. Additionally, the model boosts average machine utilization and agent utilization by 8\% each, indicating a more efficient use of resources.

\begin{table}[t]
\centering

\begin{tabular}{|c|c|c|}
\hline
Model       &   MAPPO           & AB-MAPPO              \\ \hline
Collected   & 10.2(1.0)         & \textbf{11.86 (0.31)}  \\ \hline
Delivered   & 8.74 (0.78)       & \textbf{10.49 (0.43)}  \\ \hline
Collisions  & 15.02 (0.96)      & \textbf{8.99 (1.87)}   \\ \hline
Avr(MU)     & 0.51 (0.11)       & \textbf{0.59 (0.16)}   \\ \hline
Avr(AU)     & 0.51 (0.05)       & \textbf{0.59 (0.04)}    \\ \hline

\end{tabular}
\caption{Evaluation results for our model (AB-MAPPO) compared to MAPPO: showing the Mean and (standard deviation) of the number of collected parts, delivered parts, collisions, and average machine and agent utilization.} 
\label{tab:ComparisonResults}
\end{table}

\begin{figure}[t]
  \centering
  \includegraphics[width=0.9\columnwidth]{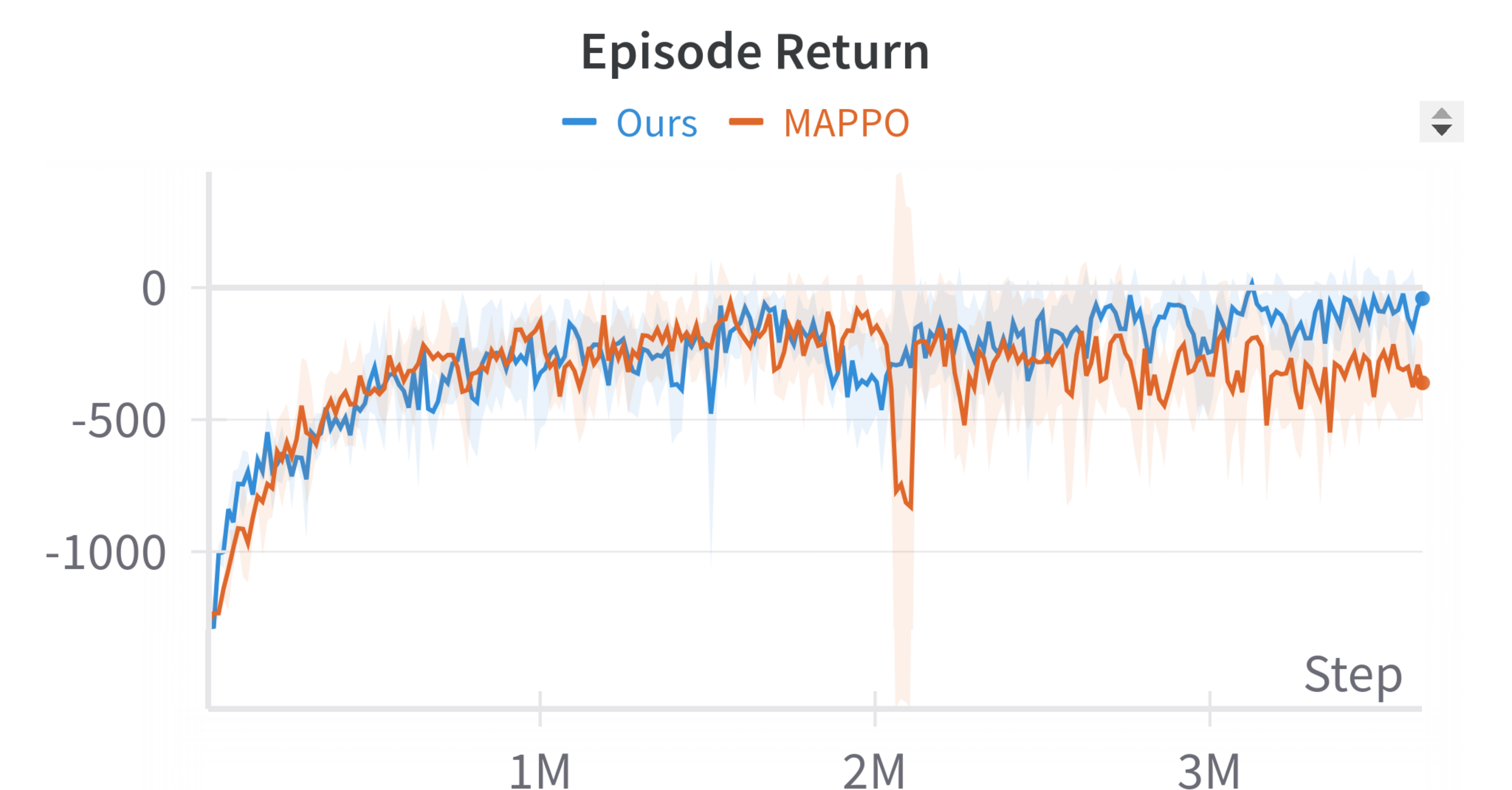}
  \caption{The total episode return for AB-MAPPO compared to MAPPO}
  \label{fig:episode_return}
\end{figure}

The total episode return for AB-MAPPO in comparison to MAPPO is shown in figure \ref{fig:episode_return}. In the beginning, the two models have similar performance, but at around 2.5 million steps ours start to outperform MAPPO and gradually continues to improve while MAPPO reaches a plateau. 

\begin{figure}[t]
  \centering
  \includegraphics[width=0.49\columnwidth]{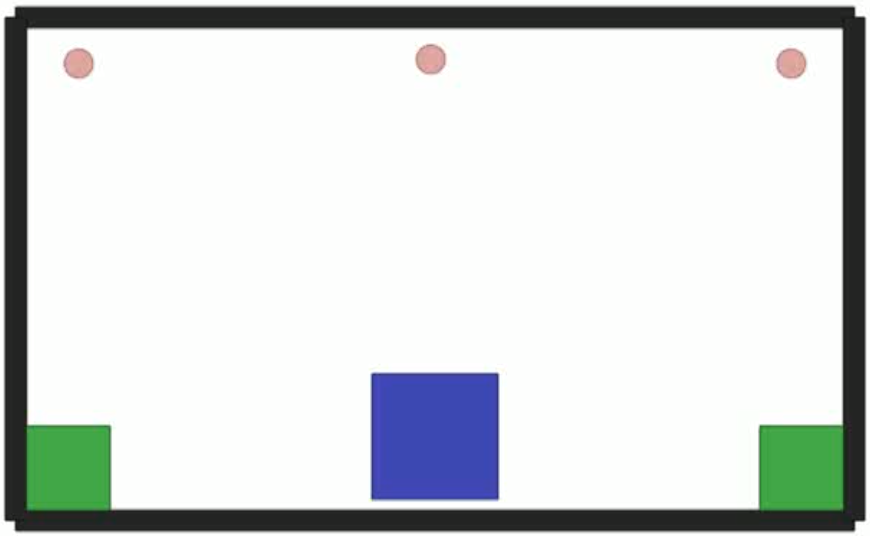}
  \includegraphics[width=0.49\columnwidth]{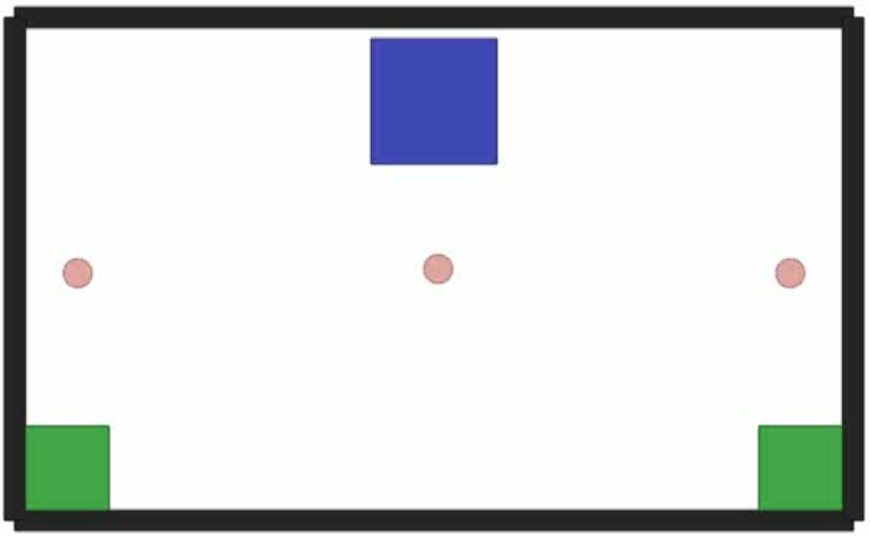}
  \includegraphics[width=0.49\columnwidth]{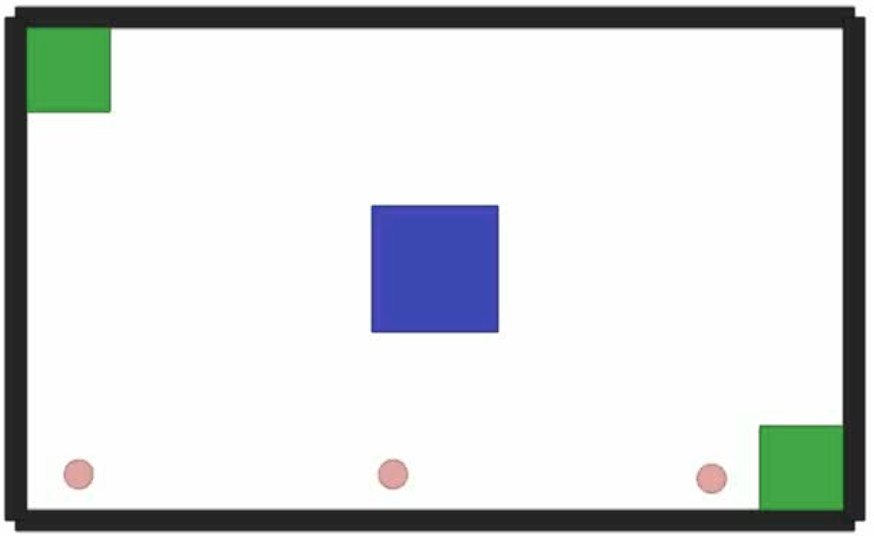}
  \includegraphics[width=0.49\columnwidth]{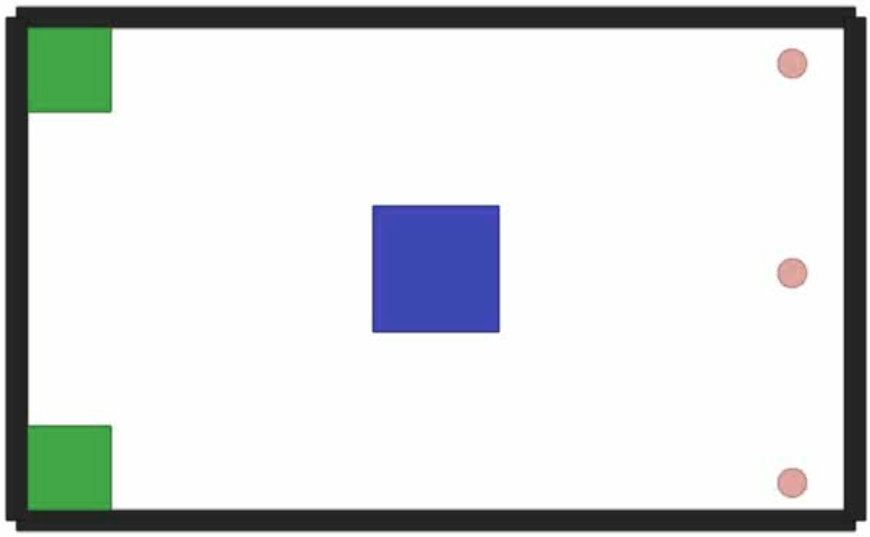}
  \caption{Examples of environment layouts with good performance}
  \label{fig:good_layouts} 
\end{figure}

\begin{figure}[t]
  \centering
  \includegraphics[width=0.49\columnwidth]{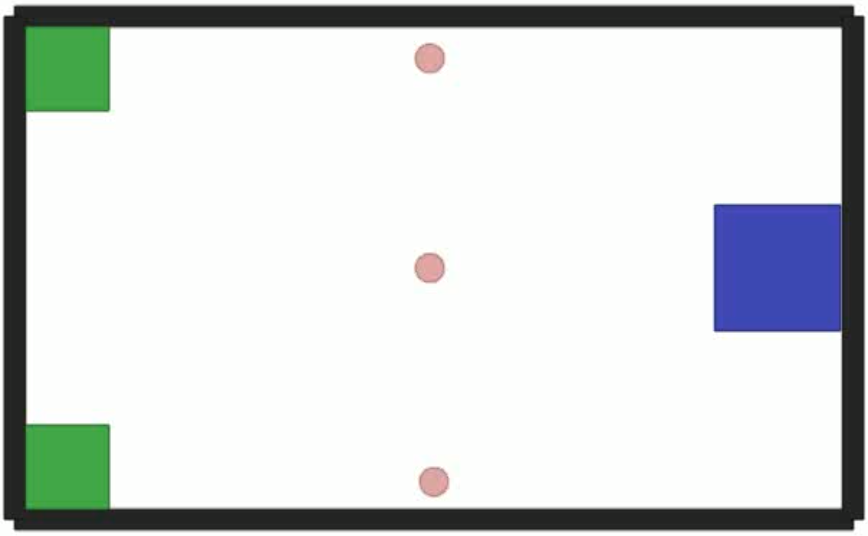}
  \includegraphics[width=0.49\columnwidth]{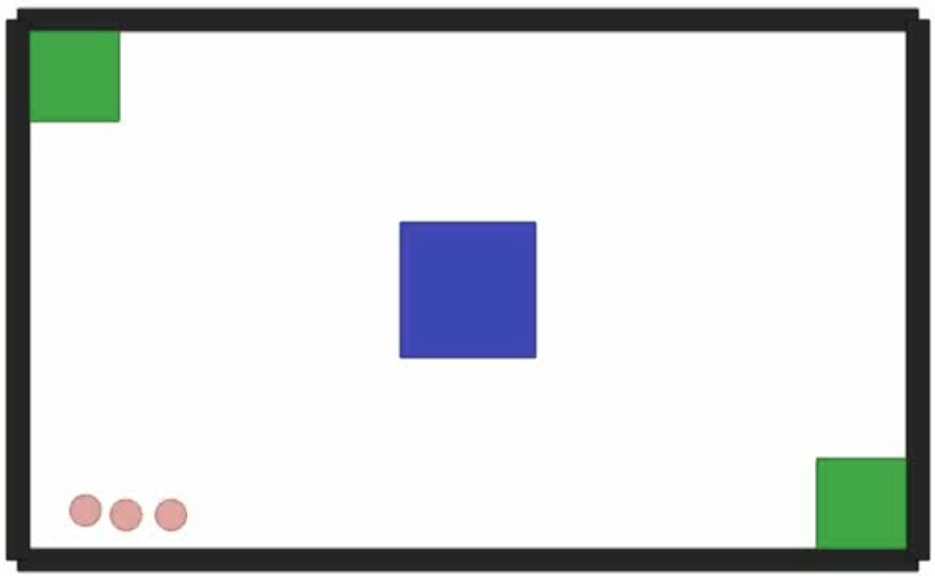}
    
  \caption{Examples of environment layouts with less optimal performance: agents learn to tend for one machine and ignore the other.}
  \label{fig:poor_layouts}
\end{figure}

To assess the adaptability of our model to various industrial settings, we conducted training and evaluation in multiple environment layouts without adjusting the hyperparameters. This approach was designed to test whether the model could be effectively deployed in different factory setups without requiring layout-specific tuning. The experiments maintained consistent dimensions for the environment, machines, agents, and storage area, varying only the layout configurations. Figure \ref{fig:good_layouts} displays examples of environment layouts where the model demonstrated robust performance. Conversely, Figure \ref{fig:poor_layouts}  highlights layouts where the model’s performance was comparatively weaker.

Empirical testing revealed that our model performs optimally when the distance between the machines and the storage area is minimized. This configuration facilitates quicker part deliveries, enhancing overall efficiency. In contrast, a larger distance from the agents' initial positions to the machines does not significantly impact performance, suggesting that initial agent placement is less critical.

However, significantly increasing the distance between the machines and the storage area adversely affects performance. We attribute this decline to the increased challenge agents face in locating the storage area after part collection, necessitating exploration over a larger area. The placement of the storage area—whether centrally or peripherally—does not markedly influence performance. Nevertheless, starting agents in close proximity to one another can impede their learning due to an increased incidence of collisions at the outset, which the collision penalty exacerbates by discouraging exploration. 


\section{Ablation Study}\label{sec:AblationStudy}

\subsection{Experiments on Observation Design}\label{subsec:ObservationAblation}

To develop an effective observation strategy for our model, we initially assessed various observation configurations using MAPPO to determine their impact on performance. Table \ref{tab:ObservationResults} presents a baseline experiment (Exp1), which incorporates all observation components detailed in \ref{subsec:Observation} including all agents' velocities and the relative positions of the walls.

Subsequent experiments each modify a single aspect of the baseline observation to isolate and measure its effect on performance: Where Exp2 adds the time elapsed since a part became ready at a machine but remained uncollected, Exp3 excludes agents' velocity data, Exp4 removes the normalization step from the observation process, Exp5 changes the representation of entities from the center point to the positions of two opposite corners, Exp6 includes the positions of machine blockers, and Exp7 omits the walls' positions from the observation). This systematic approach allows us to identify the most influential observation factors and optimize the model's input data for better learning outcomes and performance.



\begin{table*}[t]
\centering
\begin{tabular}{|c|c|c|c|c|c|}
\hline
Exp &  Collected & Delivered & Collisions & Avr(MU) & Avr(AU) \\
\hline
1 & 7.66(0.7)& 6.29(0.7)& 13.47(4.2) &  0.38(0.2)&  0.38(0.0)\\ \hline
2 & 1.26(0.1)& 0.01(0.0)& 10.25(3.3) &  0.06(0.0)&  0.06(0.0)\\ \hline
3 & 9.25(1.1)& 7.8(1.1)& 13.96(1.8) &  0.46(0.1)&  0.46(0.0)\\ \hline
4 & 7.78(1.2)& 6.24(1.3)& 8.67(2.5) &  0.39(0.2)&  0.39(0.1)\\ \hline
5 & 9.18(0.5)& 7.74(0.5)& 7.72(1.3) &  0.46(0.2)&  0.46(0.0)\\ \hline
6 & 8.15(1.0)& 6.64(1.0)& 9.22(1.9) &  0.41(0.3)&  0.41(0.0)\\ \hline
7 & 7.49(0.1)& 5.78(0.2)& 9.78(2.0) &  0.38(0.3)&  0.37(0.1)\\ \hline
\end{tabular}
\caption{Evaluation results for various observation components: showing the Mean and (standard deviation) of the number of collected parts, delivered parts, collisions, and average machine and agent utilization.} 
\label{tab:ObservationResults}
\end{table*}

Our experimental findings indicate that certain observation modifications enhance model performance while others may impede it. Specifically, the inclusion of wall and machine blocker observations, the use of two-corner representations for entities, and normalization procedures significantly improved parts delivery outcomes. Conversely, tracking agents' velocities and the duration since parts became ready without collection appeared to hinder learning. In terms of safety, the two-corner representation and normalization generally reduced the number of collisions, suggesting that more detailed information in the observation helps in avoiding mishaps. Interestingly, the inclusion of wall observations did not demonstrate a clear benefit and appeared not to affect safety positively.

Furthermore, enhancements in parts delivery correlated well with improvements in both machine and agent utilization rates. This suggests that factors contributing positively to operational efficiency also tend to optimize resource usage.

\subsection{Experiments on Reward Design}

This section continues our evaluation by focusing on the different components of the reward structure to determine their individual impacts on performance. Utilizing the same base experiment setup as before, all reward components outlined in \eqref{equ:RT} were initially included, along with reward sharing where all agents receive the pick or place reward if at least one agent successfully picks up or places a part, respectively. Subsequent experiments each isolate and modify a specific reward feature to analyze its effects. Where Exp2 eliminates the time penalty, Exp3 removes reward sharing, Exp4 omits the uncollected parts penalty, Exp5 applies an increasing uncollected parts penalty as described in the latter part of \eqref{equ:RU}, and Exp6 removes the distance-based reward. Table \ref{tab:RewardResults} details these comparisons.


\begin{table*}[t]
\centering
\begin{tabular}{|c|c|c|c|c|c|}
\hline
Exp &  Collected & Delivered & Collisions & Avr(MU) & Avr(AU) \\
\hline
1 & 7.66(0.7)& 6.29(0.7)& 13.47(4.2) &  0.38(0.2)&  0.38(0.0)\\ \hline
2 & 6.72(1.8)& 5.4(1.9)& 18.33(9.0) &  0.34(0.0)&  0.34(0.0)\\ \hline
3 & 8.44(0.7)& 7.39(0.5)& 27.25(9.8) &  0.42(0.2)&  0.42(0.1)\\ \hline
4 & 4.79(0.6)& 3.46(0.2)& 6.58(2.1) &  0.24(0.1)&  0.24(0.0)\\ \hline
5 & 1.55(0.1)& 0.28(0.2)& 167.48(53.7) &  0.08(0.0)&  0.08(0.0)\\ \hline
6 & 6.57(2.6)& 4.99(3.4)& 7.57(4.6) &  0.33(0.2)&  0.33(0.1)\\ \hline
\end{tabular}
\caption{Evaluation results for various reward components: showing the Mean and (standard deviation) of the number of collected parts, delivered parts, collisions, and average machine and agent utilization.} 
\label{tab:RewardResults}
\end{table*}

Generally, the removal of any reward component negatively impacted productivity and resource utilization. Notably, not sharing rewards and applying a fixed penalty for uncollected parts yielded better results. From a safety perspective, fewer reward terms led to increased safety, except in the case of the time penalty. Sharing rewards and using fixed penalties for uncollected parts also resulted in fewer collisions.

Additionally, we assessed the performance of combining the most effective observation and reward options. Table  \ref{tab:CombinationResults} presents these results. Not all combinations led to performance enhancements. For instance, Exp1, which combines removing velocity observation (RV) with observing machine blockers' positions (OMB), did not improve outcomes. However, Exp2, which combines RV with no reward sharing (NSH), showed enhanced performance, as did Exp3, which pairs no velocity observation (NV) with two-corner entity representation (COR). Contrarily, combining all selected options (NV, COR, OMB, and NSH) in Exp4 did not yield better results. Ultimately, the setup from Exp2 was chosen as the optimal configuration for our observation and reward strategy.


\begin{table*}[t]
\centering
\begin{tabular}{|c|c|c|c|c|c|}
\hline
Exp &  Collected & Delivered & Collisions & Avr(MU) & Avr(AU) \\
\hline
1  & 9.03(1.4)& 7.56(1.5)& 11.59(1.1) &  0.45(0.2)&  0.45(0.1)\\ \hline
2 & \textbf{10.2(1.0)} & \textbf{8.74(0.8)}& 15.02(1.0) &  \textbf{0.51(0.1)}&  \textbf{0.51(0.1)}\\ \hline
3 & 9.8(0.7)& 8.16(1.0)& \textbf{7.64(2.0)} &  0.49(0.2)&  0.49(0.1)\\ \hline
4 & 8.51(0.6)& 7.43(0.5)& 21.26(1.6) &  0.43(0.4)&  0.43(0.1)\\ \hline %
\end{tabular}
\caption{Combining the best performing options: Mean (stddev).} \label{tab:CombinationResults}
\end{table*}

\section{Conclusion}
This study applies MARL to a challenging real-world industrial scenario (machine tending) aiming to leverage the full potential of multiple mobile robots. We enhanced the MAPPO architecture with AB-MAPPO, incorporating a novel attention-based encoding to improve feature representation significantly. These innovations establish a new baseline and pave the way for future advancements. Rigorously tested across diverse environments and supported by a comprehensive ablation study, AB-MAPPO demonstrated its efficacy and adaptability. Looking forward, we plan to extend these innovations to more dynamic scenarios, including real-time material feeding and interactive object manipulation, to bridge the gap between theoretical research and practical applications.

\bibliography{main}

\end{document}